\pdfoutput=1
\documentclass[a4paper]{article}
\usepackage{spconf,amsmath,graphicx,booktabs,amssymb,subfigure,threeparttable,multirow,float,hyperref,url,color}
\hypersetup{
    colorlinks=true,
    linkcolor=red,
    filecolor=red,      
    urlcolor=red,
    citecolor=green,
}

\title{Dive into Big Model Training}
\name{Qinghua Liu$^1$$^,$$^2$, Yuxiang Jiang$^1$}
\address{$^1$JD.com Inc, Beijing, China \\
$^2$Tianjin University, Tianjin, China}
\begin{document}
\ninept

\maketitle

\begin{abstract}
% The increasing scale of model size and continuous improvement of performance heralds the arrival of the Big Model era. This report combines algorithmic and engineering perspectives on Big Model training. In this report, we first elicit the need for distributed training by reviewing the development of large models and training algorithms. We summarize the existing methods for big model training into three main categories: training parallelism, memory saving technologies, and mixture-of-expert. According to the dimension of parallelism that takes place, training parallelism can be further categorized into data, pipeline, and tensor parallelism. Memory-saving technologies are orthogonal and complementary to parallelism. And mixture-of-expert algorithm scales up the model size with a constant computational cost. A continuously updated paper list  of big model training is provided$\footnote{https://github.com/qhliu26/BM-Training}$.
The increasing scale of model size and continuous improvement of performance herald the arrival of the Big Model era. In this report, we explore what and how the big model training works by diving into training objectives and training methodologies. Specifically,training objectives describe how to leverage web-scale data to develop extremely capable and incredibly large models based on self-supervised learning, and training methodologies which are based on distributed training describe how to make big model training a reality. We summarize the existing training methodologies into three main categories: training parallelism, memory-saving technologies, and model sparsity design. Training parallelism can be categorized into data, pipeline, and tensor parallelism according to the dimension of parallelism that takes place. Memory-saving technologies are orthogonal and complementary to training parallelism. And model sparsity design further scales up the model size with a constant computational cost.
A continuously updated paper list of big model training is provided$\footnote{https://github.com/qhliu26/BM-Training}$.
\end{abstract}

\noindent\textbf{Index Terms}: Deep Learning, Distributed Training, Training Parallelism

\section{Introduction}
\label{sec:intro}
The past several years have witnessed the scale of the model increase as datasets become larger and large-scale computation becomes more available. The rise of self-supervised learning algorithms has enabled models to unleash the power of extensive unlabeled datasets and become the carriers of intelligence. The prosperous training methodologies, including training parallelism and memory-saving technologies, make it possible to train outrageously large neural networks on many GPUs in an efficient manner.
The term \textbf{Big Model} refers to a big data-driven and multi-domain capable model with a large number of parameters. And the \textbf{training} objective describes how to leverage broad data to transform deep neural network architecture into a big model with the help of large-scale computing power.

\subsection{Development of Big Model Scale}
\label{ssec:develop}
The number of parameters of deep neural networks has been growing rapidly in recent years. As shown in Table~\ref{tb:model}, the model size grows rapidly over time. Since the advent of the GPT series, the era of Big Model has begun.

\begin{table}[htb]
\centering
\caption{Large-scale models released in recent years.}
\begin{tabular}{lcc}
\hline
\textbf{Name} & \textbf{\#Param} & \textbf{Date} \\ \hline
GPT~\cite{radford2018improving} & 110 million & Jun 2018 \\
BERT~\cite{devlin2018bert} & 340 million & Oct 2018\\
GPT-2~\cite{radford2019language}  & 1.5 billion &  Feb 2019 \\
Megatron-LM~\cite{shoeybi2019megatron} & 8.3 billion &  Sep 2019 \\
Turing-NLG~\cite{rajbhandari2020zero} & 17 billion &  Feb 2020 \\
GPT-3~\cite{brown2020language} & 175 billion &  May 2020 \\
Switch Transformer~\cite{fedus2021switch} & 1.6 trillion &  Jan 2021 \\
BaGuaLu~\cite{ma2022bagualu} & 174 trillion &  April 2021 \\
Megatron-Turing NLG~\cite{smith2022using} & 530 billion &  Jan 2022 \\
PaLM~\cite{chowdhery2022palm} & 540 billion &  April 2022 \\
\hline
\end{tabular}
\label{tb:model}
\end{table}

According to~\cite{shoeybi2019megatron}, Megatron-LM uses 32 DGX-2H servers (a total of 512 Tesla V100 GPUs) to scale transformer-based models up to 8.3 billion parameters based on tensor parallelism (Sec~\ref{ssec:tp}). ZeRO (Sec~\ref{ssec:zero}) proposed by Rajbhandari et al. in~\cite{rajbhandari2020zero} could fit more than 1 trillion parameters on 1024 GPUs with model and data parallelism. However, it is only on a theoretical level and due to resource constraints, they actually train a 17B language model with state-of-the-art performance and the largest model size by that time. The first \textit{Trillion}-scale parameter model Switch Transformer~\cite{fedus2021switch} which is based on the mixture-of-experts algorithm (Sec~\ref{sec:sparsity}) has come out in 2021. In the same year, the number of parameters in Bagualu~\cite{ma2022bagualu} reached a record-breaking 174 trillion, which even rivals the number of synapses in a human brain.

\subsection{Why Big Model Prospers}
\label{ssec:why}
The upper limit of the parameter scale has been broken over the past few years along with the state-of-the-art performance. The reasons for the continued interest in Big Model include the following:

\textbf{Big Model performs better.}
Intuitively, the best performers on the standard dataset are basically pre-trained Big Models. By scaling up the model size, big models benefit from significant performance improvements across a wide range of machine learning problems~\cite{liu2022swin, sauer2022stylegan, radford2019language, shoeybi2019megatron}. 
Kaplan et al.~\cite{kaplan2020scaling} from OpenAI demonstrate exhaustively that model performance depends strongly on scale and has a power-law relationship with the number of parameters, size of the dataset, and amount of computing. This study further suggests that larger models will continue to perform better.

\textbf{Big Model is more capable.}
The Big Model shows the ability to adapt to multiple downstream tasks after being pretrained on a large amount of data. The multiple downstream tasks and data requirements are trivial and long-tailed distribution. Standard pre-training and fine-tuning paradigm~\cite{bengio2006greedy} generally require updating the entire model weight. As we scale the parameter size from GPT-2 (1.5B) to GPT-3 (175B), GPT-3 becomes a few-shot learner which can better achieve downstream task adaptation without fine-tuning. In particular, it can be adapted via natural language prompts to do a passable job on a wide range of tasks despite not being trained explicitly to do many of those tasks.

\subsection{Challenges in Big Model training}
\label{ssec:challenge}

\begin{table}[htb]
\centering
\caption{Estimated training time on one NVIDIA V100 GPU.}
\begin{tabular}{lccc}
\hline
\textbf{Name} & \textbf{\#Param} & \textbf{Dataset Size}  & \textbf{Time}\\ \hline
GPT~\cite{radford2018improving} & 110M & 4GB & 3 days\\
BERT~\cite{devlin2018bert} & 340M & 16GB & 50 days\\
GPT-2~\cite{radford2019language}  & 1.5B & 40GB & 200 days\\
GPT-3~\cite{brown2020language} & 175B & 560GB & 90 years \\
\hline
\end{tabular}
\label{tb:challenge}
\end{table}

The significant performance improvement by bigger models comes with many practical challenges. The bottleneck restricting our efficient training of big models mainly lies in computing resources and network communication. 
Big Model training is challenging for the following reasons: 
\begin{enumerate}
    \item as shown in Table~\ref{tb:challenge}, the number of computing operations required results in unrealistically long training time;
    \item it is no longer possible to fit the entire model on a single accelerator (mainly GPU and TPU) due to limited memory capacity;
    \item computational scalability is not linear due to the communication cost and memory redundancy, for example, the training time may not be reduced by a factor of two if we use twice the number of GPUs.
\end{enumerate}

\subsection{Overview of this report}
\label{ssec:overview}
This report is organized as follows. 
In Section~\ref{sec:ssl}, we review training algorithms for Big Model based on self-supervised learning.
In Section~\ref{sec:training}, we introduce distributed training needed for Big Model training.
In Section~\ref{sec:parallelism}, we introduce three kinds of training parallelism methods including DP (Data Parallelism), PP (Pipeline Parallelism) and TP (Tensor Parallelism). 
In Section~\ref{sec:mem}, we introduce some widely used memory-saving technologies.
In Section~\ref{sec:sparsity}, we introduce the sparsely gated mixture-of-experts approach which greatly enlarges model size with the same amount of computing resources.
Finally, in Section~\ref{sec:discussion} and~\ref{sec:conclusion}, we bring up a discussion on Big Model and conclude the report.

\section{Training Objectives}
\label{sec:ssl}
Before diving into big model training methodologies, this section first introduces training objectives~\cite{bommasani2021opportunities} of transforming a model architecture and a large amount of data into a Big Model based on self-supervised learning (SSL).

The traditional paradigm for training deep neural networks is supervised learning, and the performance of the model grows roughly logarithmically with the size of the annotated dataset~\cite{sun2017revisiting}. However, rare and expensive data with annotation quickly becomes a scalability bottleneck for continuous improvement of state-of-the-art model performance. Hence, it stimulates research on self-supervised learning in academia. Over the past few years, SSL has achieved great success in the fields of natural language processing~\cite{radford2018improving,devlin2018bert}, computer vision~\cite{dosovitskiy2020image,liu2021swin} and speech processing~\cite{hsu2021hubert, chen2021wavlm}. It leverages large amounts of data to learn universal representations, which can benefit almost all downstream tasks.

The term "self-supervised learning" was first introduced in robotics. Then the machine learning community further develops this idea and describes it as "the machine predicts any part of its input for any observed part"~\cite{bengio2021deep}. Current SSL methods can be categorized into generative, contrastive, and generative-contrastive according to their objectives~\cite{liu2021self}. We enumerate below their respective definitions and notable works that use the corresponding algorithms:
\begin{enumerate}
    \item \textbf{Generative:} train an encoder to encode input into an explicit vector and a decoder to reconstruct the input from the vector, either continuous or discrete. SSL methods based on generative models include autoregressive models (GPT~\cite{radford2018improving}, PixelRNN~\cite{van2016pixel}, WaveNet~\cite{oord2016wavenet}), auto-encoding models (BERT~\cite{devlin2018bert}, MAE~\cite{he2022masked}, VQ-VAE~\cite{van2017neural}) and hybrid generative models (XLNet~\cite{yang2019xlnet}).
    \item \textbf{Contrastive:} train an encoder to encode input into an explicit vector to measure similarity. The well-known works include CPC~\cite{oord2018representation} which uses the contrastive InfoNCE loss to discriminate the correlated positive samples from negative context samples, and MoCo~\cite{he2020momentum} which leverages instance discrimination via momentum contrast.
    \item \textbf{Generative-Contrastive:} train an encoder-decoder to generate fake samples and a discriminator to distinguish them from real samples. It leverages the discriminative loss function as the objective. The most influential work is Generative Adversarial Networks (GAN)~\cite{goodfellow2014generative} and there are many of its variants~\cite{radford2015unsupervised,iizuka2017globally,karras2019style}.
\end{enumerate}

The pre-training and fine-tuning paradigm~\cite{bengio2006greedy} is the most widely adopted pipeline for the SSL method to transfer to the downstream task. Specifically, researchers firstly design elaborate pretext tasks (e.g. Next Sentence Prediction task in BERT~\cite{devlin2018bert}) to help models learn critical features from a large number of unlabeled datasets. Then they can fine-tune the pretrained model on a variety of target labeled datasets without training from scratch. Moreover, the recent GPT-3~\cite{brown2020language} has shown that large-scale SSL models can achieve competitive performance via few-shot adaptation instead of full finetuning, which stimulates researchers' desire to explore the potential of bigger models. However, Big Model always brings expensive training costs and unbearable training time. The following parts will introduce some relevant system designs and software platforms to democratize Big Model training.

\section{Distributed Training}
\label{sec:training}
Although Big Model brings better performance and more capabilities, the training of Big Model is not just an algorithmic challenge but a systemic challenge. A GPT-3 with 175B parameters can never be fit into one GPU's memory even with the most powerful GPU. Distributed computing can't be avoided in Big Model training.

If the Big Model is like a vehicle, the distributed training is the highway that makes the model run faster. As the cornerstone of parallel processing of big data and big model training, distributed training consists of two parts: the training framework and the distributed system. Distributed training makes large models go from untrainable to trainable, from unacceptable training time and money consumption to convergent cost consumption. Distributed training based on backward pass usually has the following main processes: slice data loading, forward pass, backward pass, set communication, and parameter update.

\subsection{Training Framework}
\label{ssec:framework}
The main process of large model training is usually built through the training framework. Based on the framework, researchers build the model layer, combined the layer to forward process, load training data and pre-processing, construct the loss function and validation process.

The most famous early training framework is Caffe~\cite{jia2014caffe}, which requires users to write backward propagation code in addition to forward code. Moreover, the original Caffe does not support multi-node distributed training. After subsequent development, it supports multi-machine training based on the MPI~\cite{open2013high} communication framework. However, there are few users of distributed training based on Caffe. The current mainstream distributed training frameworks are attracted by more than 90\% of developers from two modern framework, namely Pytorch~\cite{paszke2019pytorch} and Tensorflow~\cite{abadi2016tensorflow}. These modern frameworks allow researchers to focus on the forward process, optimizer efficiency, and loss design while using automatic differentiation techniques to implicitly express the computation of the gradient compute and backward process. The most important feature of the modern training framework for Big Model is the multi-node multi-gpu setting which reduces the time to configure distributed GPUs with less expert knowledge.

\subsection{Distributed system}
\label{ssec:system}

\begin{figure}[htb]
	\centering
	\includegraphics[width=\linewidth]{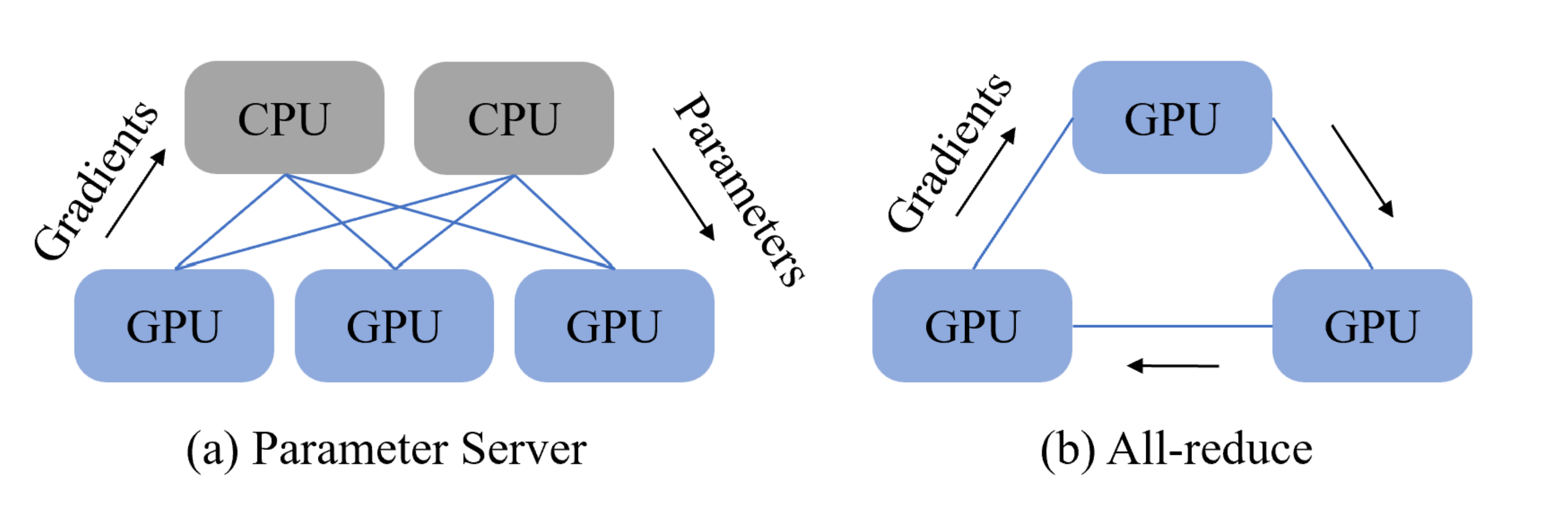}
	\caption{Illustration of Parameter Server (a) and All-reduce (b) architecture.}
	\label{fig:reduce_ps}
\end{figure}

There are two main categories of distributed training architectures: Parameter Server (PS)~\cite{li2014communication} and All-reduce~\cite{patarasuk2009bandwidth} as is shown in Fig~\ref{fig:reduce_ps}.

The PS architecture typically contains \textit{workers} and \textit{PS}. Worker usually run on GPUs to perform forward and backward pass and push the gradients to PS on CPUs to update the parameters. Then workers pull the latest parameters from PS and start the next iteration. However, communication workload and network congestion become the performance bottleneck when scaling PS to larger clusters.

In all-reduce, only GPU machines are involved. GPUs compute the gradients of the model parameters independently and then aggregate gradients based on collective communication method. Specifically, a Ring all-reduce operation can be divided into a \textit{reduce-scatter} and an \textit{all-gather} operation. Reduce-scatter partitions $M$ bytes gradients into $n$ parts and each node send $\frac{(n-1)M}{n}$ traffic to other nodes.
Then all-gather requires each node to broadcast its reduced
part to all other $(n-1)$ nodes to obtain the result in an iteration. As a decentralized and homogeneous architecture, all-reduce achieves higher performance than PS in most distributed training tasks. However, it can't utilize extra non-worker nodes and further research has shown that it is no longer optimal with additional CPU machines~\cite{jiang2020unified}.

As a more general system for distributed training, BytePS~\cite{jiang2020unified} can leverage spare CPU and bandwidth resources in the cluster to accelerate DNN training. It offers a framework for communication that is demonstrated to be optimal in heterogeneous GPU/CPU clusters, and current all-reduce and PS can be regarded as two specific cases of BytePS.

\section{Training Parallelism}
\label{sec:parallelism}
As the scale of the model and dataset increases, it becomes hard to deploy training tasks on an individual GPU machine. Even if we could fit the entire model on one GPU, the training time would become a nightmare. Thus, it is natural to employ distributed training methods and multi-processes to reduce the training burden. The training algorithms can be categorized into several popular types, such as synchronous vs asynchronous, cross-iteration vs intra-iteration, and data parallel vs model parallel~\cite{li2020pytorch}.

In this report, we categorize these techniques according to the dimension of parallelism that takes place, including data (batch size), model (forward and backward pass), and tensor (matrix operations)~\cite{weng2021large}.

\subsection{Data Parallelism}
\label{ssec:dp}
As the simplest and most common way to scale training, Data Parallelism replicates the same model weights to multiple devices and distributes a portion of the data to each device for simultaneous processing, which is equivalent to parallelizing the training process along the \textit{batch} dimension. 
In particular, each process performs forward and backward propagation on different subsets of data samples and synchronizes either gradients or updated parameters depending on the algorithm.
There are three main synchronization approaches:
\begin{enumerate}
    \item \textbf{Bulk Synchronous Parallel} (BSP)~\cite{valiant1990bridging}: Workers wait for each other to synchronize at the end of every iteration, which reduces the staleness of weights used to compute gradients and ensures good statistical efficiency. However, waiting for gradients from other workers significantly decreases hardware efficiency and shifts the training bottleneck to communication when the model scales. 
    \item \textbf{Asynchronous Parallel} (ASP)~\cite{ahmed2012scalable}: Each worker proceeds with the computation for the next iteration without waiting for gradients from other workers, which reduces idle time but may lead to gradients being computed on stale weights and lowers statistical efficiency. Experimental studies demonstrate that ASP does not reduce training time~\cite{chen2016revisiting}.
    \item \textbf{Stale Synchronous Parallel} (SSP)~\cite{ho2013more}: When a worker asks for updated parameter $\theta$, the model will give it a stale version that excludes recent gradient updates $\delta$. Specifically, a worker reading $\theta$ at iteration $c$ gets all $\theta$ accumulated from iteration 0 to $c-s-1$, where $s \geq 0$ is a user-controlled threshold. In the SSP model, workers can perform more computations instead of waiting for other workers to finish.
\end{enumerate}

Training frameworks (Sec~\ref{ssec:framework}) provide native distributed Data Parallelism APIs for developers.

For example, PyTorch offers \textit{DataParallel} (DP) for single-process multi-thread data parallel training using multiple GPUs on the same machine, and \textit{DistributedDataParallel} (DDP)~\cite{li2020pytorch} for multi-process data parallel training across GPUs and machines. DDP has provided several techniques to accelerate training since PyTorch v1.5, including bucketing gradients (which buckets multiple gradients into one AllReduce operation instead of launching an AllReduce immediately when each gradient tensor becomes available), overlapping computation with communication (which starts an AllReduce operation on gradients before the local backward pass finishes), and gradient accumulation (which conducts $n$ local training iterations before synchronizing gradients globally instead of launching AllReduce in every iteration).

\subsection{Pipeline Parallelism}
\label{ssec:pp}
\begin{figure}[htb]
	\centering
	\includegraphics[width=\linewidth]{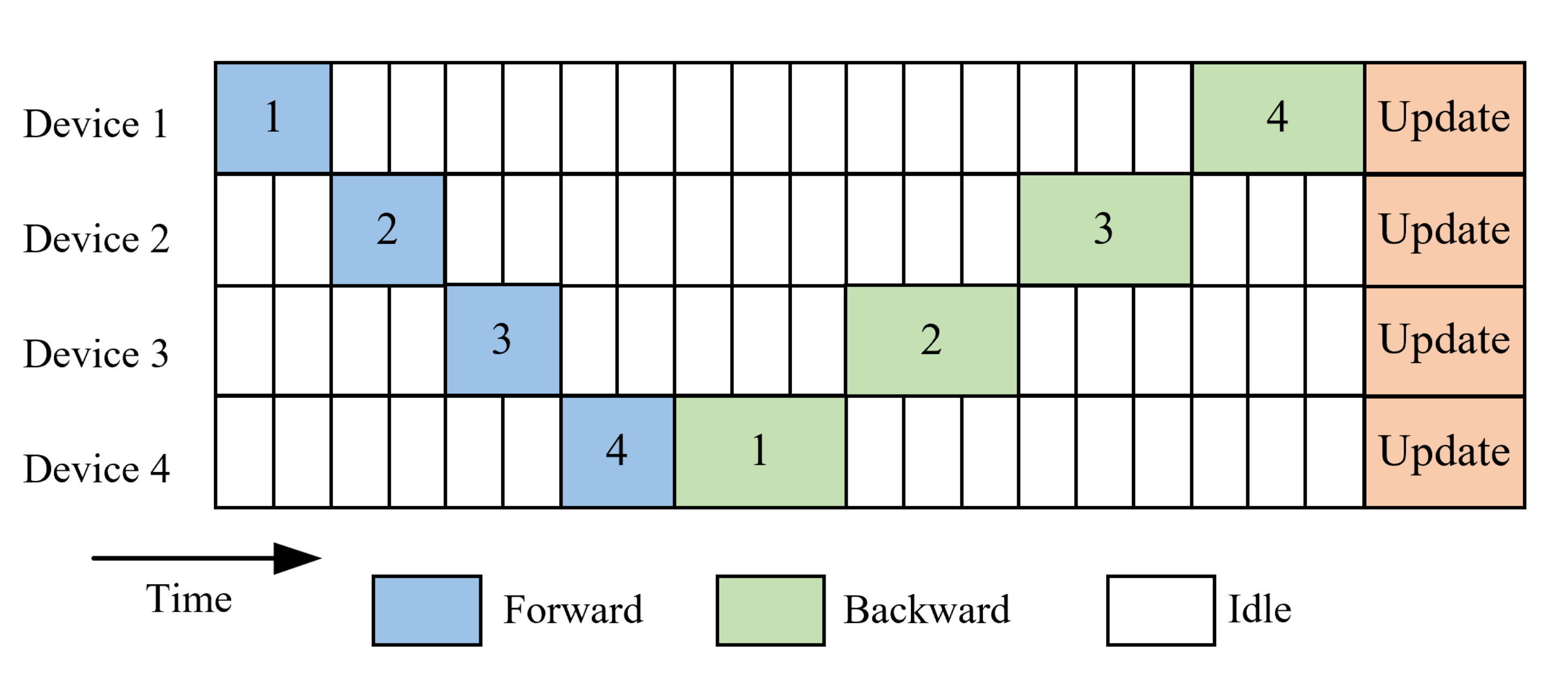}
	\caption{Model parallel training with 4 devices. Numbers indicate minibatch ID.}
	\label{fig:mp}
\end{figure}

When a model gets too big to fit on an individual GPU, it feels straightforward to partition the model across multiple GPUs, with each GPU responsible for only a portion of the model. However, a naive implementation of model parallelism results in severe under-utilization of GPU resources, as illustrated in Figure~\ref{fig:mp}. For each minibatch, there is only one active stage. When one device starts the processing stage, all other devices get stuck in a bubble of idle time. 

Pipeline Parallelism is a method to speed up neural network training by combining model parallelism with data pipelining.
The main idea is that we split one minibatch into multiple microbatches to enable devices to process several microbatches simultaneously. During the forward pass, each device sends intermediate activations to the next stage. During the backward pass, each device sends the gradients of the input tensor back to the previous pipeline stage.

There are several possible ways of scheduling forward and backward microbatches across devices, either in a synchronous way such as GPipe~\cite{huang2019gpipe} or in an asynchronous way such as PipeDream~\cite{harlap2018pipedream}.

\begin{figure}[htb]
	\centering
	\includegraphics[width=\linewidth]{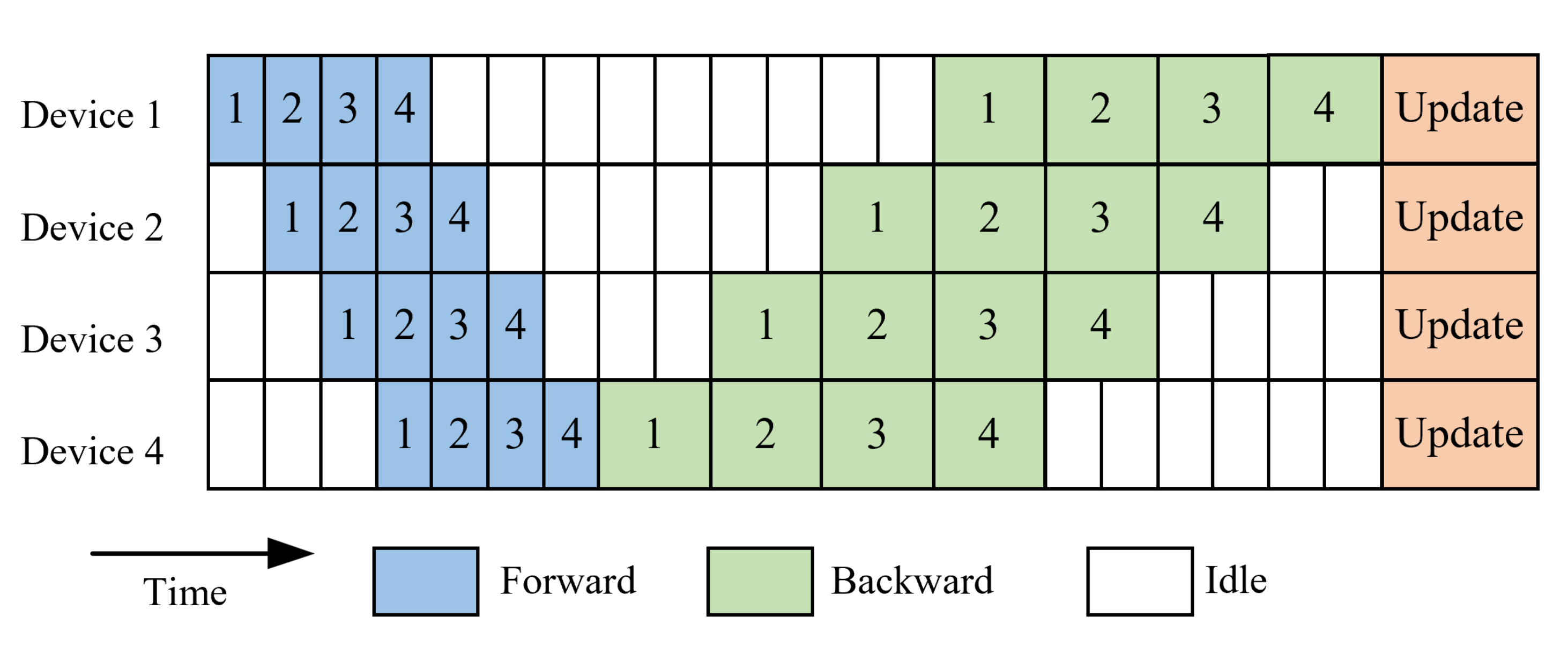}
	\caption{Illustration of Pipeline Parallelism in GPipe with 4 devices and 4 microbatches. Image based on:~\cite{narayanan2021efficient}.}
	\label{fig:gpipe}
\end{figure}

GPipe proposes a schedule where the forward passes for all microbatches in a batch are first executed, followed by backward passes for all microbatches as illustrated in Figure~\ref{fig:gpipe}.
Following~\cite{narayanan2021efficient}, here we refer to the idle time introduced by partitioning as pipeline bubble $t_{pb}$, the number of microbatches in a batch as $m$, and the number of devices used for pipeline parallelism as pipeline stages $p$.
The total amount of time spent in the pipeline bubble is $t_{pb}=(p-1)(t_f+t_b)$, where $t_f$ and $t_b$ denote the time to execute a single microbatch’s forward and backward pass. And the ideal processing time for a batch is $t_{ideal}=m(t_f+t_b)$.
Therefore, the bubble time fraction is $\frac{t_{pb}}{t_{ideal}}=\frac{p-1}{m}$.

To make the fraction as small as possible, we need $m \gg p$. The experiments in~\cite{huang2019gpipe} demonstrate that the bubble overhead is negligible under the condition that $m \geq 4 \times p$ when activation checkpointing (Sec~\ref{ssec:checkpointing}) is applied.

PipeDream schedules each worker to alternatively perform the forward and backward passes (1F1B for short), which is more memory-efficient than GPipe.
In this schedule, the devices first enter a startup state where they perform a different number of forward passes.
As soon as the output stage completes the forward pass for the first minibatch, it performs the backward pass for the same batch and then starts alternating between forward and backward passes in the following minibatches. After the startup state, each device then enters a steady state where no GPU is idle.

\begin{figure}[htb]
	\centering
	\includegraphics[width=\linewidth]{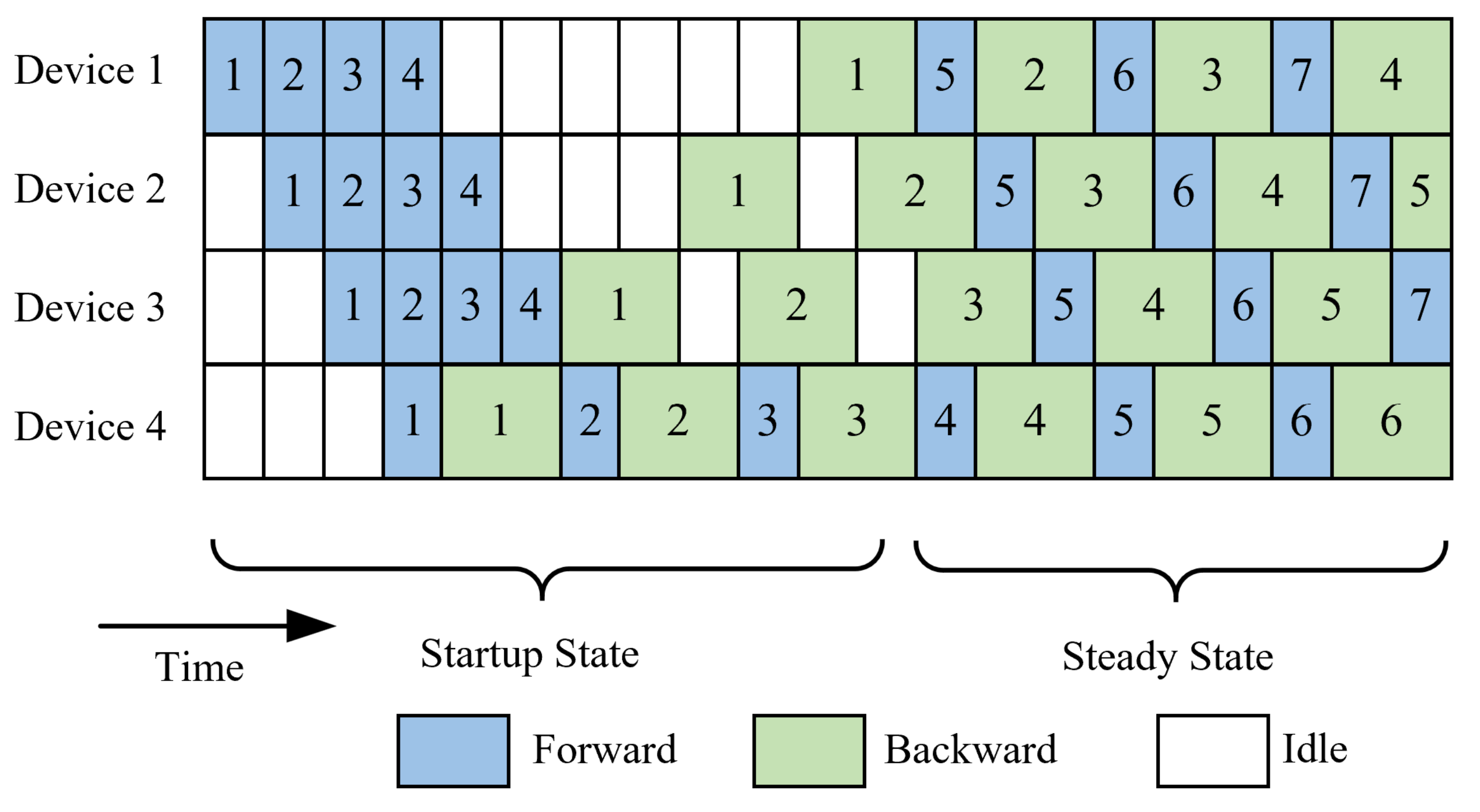}
	\caption{Illustration of 1F1B microbatch scheduling in PipeDream. Image based on:~\cite{harlap2018pipedream}.}
	\label{fig:pipedream}
\end{figure}

As we can see, a naive implementation of pipelining may lead to weight version mismatches between forward and backward passes. In GPipe (Figure~\ref{fig:gpipe}), weight gradients are accumulated and updated at the end of the minibatch. It maintains a single weight version but requires periodic sync semantics. However, in PipeDream (Figure~\ref{fig:pipedream}) which does not have end-of-batch gradient synchronization, it's likely that the naive 1F1B schedule would cause serious weight version mismatches. Specifically, when weight updates are immediately applied to the latest weight version, gradient computations may be incorrect for the forward and backward of one microbatch using a different version of weights. PipeDream, therefore, adopts a weight stashing scheme to tackle this issue. The total number of weight versions stashed is $p$ in the worst case, which is expensive for big models. PipeDream-flush~\cite{narayanan2021memory} introduces a periodical global synchronization scheme that reduces the memory footprint by sacrificing a little throughput. The number of outstanding forward passes is at most the number of pipeline stages for this schedule, but the time spent in the bubble stays the same.

\begin{figure}[htb]
	\centering
	\includegraphics[width=\linewidth]{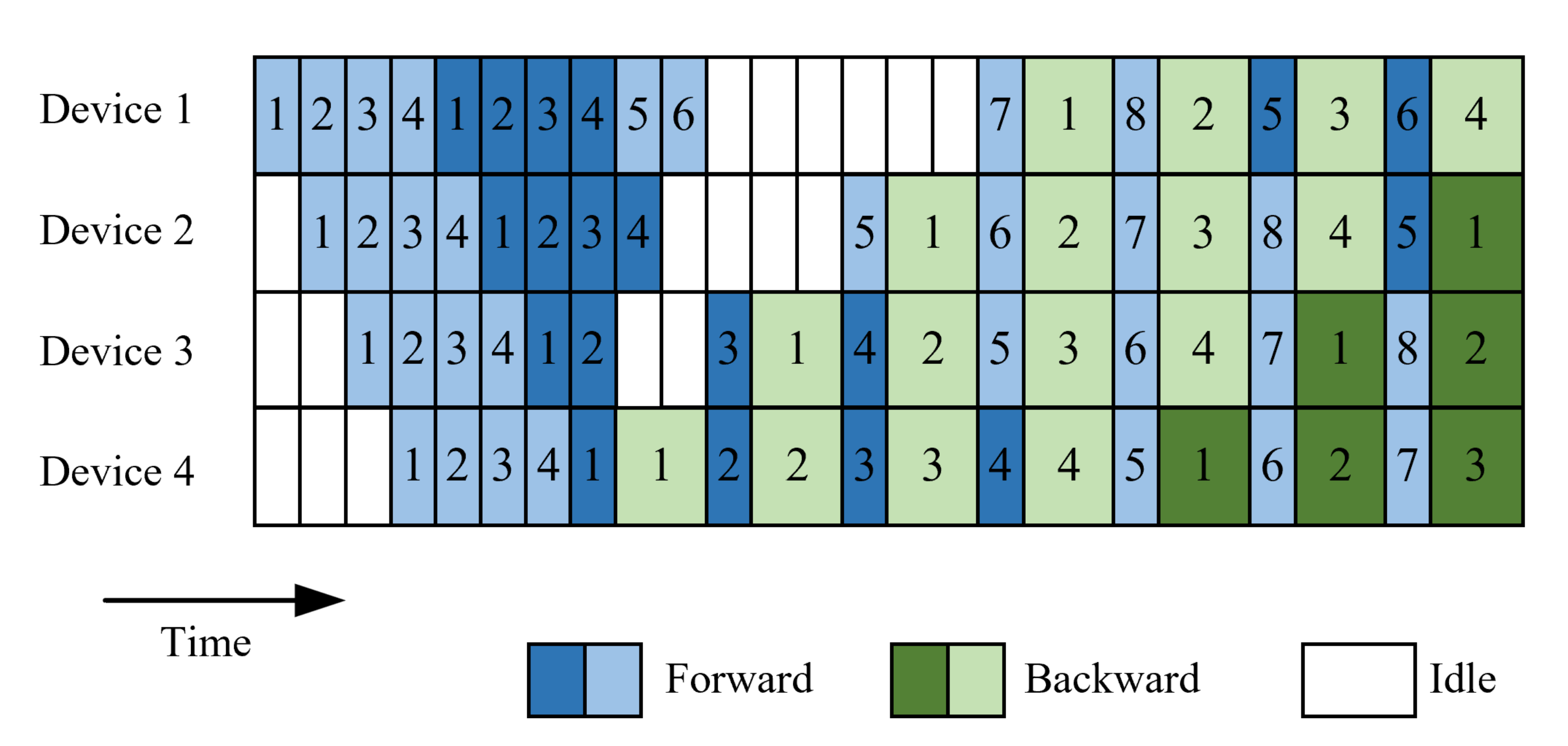}
	\caption{Illustration of interleaved 1F1B pipeline schedule. Each device is assigned 2 chunks. Light colors show the first chunk and dark colors show the second chunk. Image based on:~\cite{narayanan2021efficient}.}
	\label{fig:interleaved}
\end{figure}

To further reduce the pipeline bubble, in the interleaved 1F1B schedule~\cite{narayanan2021efficient}, each device is assigned multiple pipeline stages and performs computation on model chunks as shown in Figure~\ref{fig:interleaved}.
The number of microbatches $m$ must be an integer multiple of the pipeline stages $p$ in this schedule.
If each device has $c$ model chunks, the forward and backward time for a microbatch for each chunk reduces to $t_f/c$ and $t_b/c$. The bubble time fraction is then $\frac{t_{pb}}{t_{ideal}}=\frac{1}{c} \frac{p-1}{m}$, indicating that the bubble time is reduced by $c$. However, the bubble time reduction comes along with an extra amount of communication by $c$ times.

\subsection{Tensor Parallelism}
\label{ssec:tp}
Tensor Parallelism refers to partitioning a tensor into several parts along a specific dimension and computing them separately on different devices. It is orthogonal and complementary to Data Parallelism(Sec~\ref{ssec:dp}) and Pipeline Parallelism(Sec~\ref{ssec:pp}).

Firstly, this report introduces the implementation of 1D Tensor Parallelism used by Megatron-LM~\cite{shoeybi2019megatron} for transformer~\cite{vaswani2017attention} layers, which consist of a self-attention block followed by a two-layer multi-layer perceptron (MLP).

\begin{figure}[htb]
	\centering
	\includegraphics[width=\linewidth]{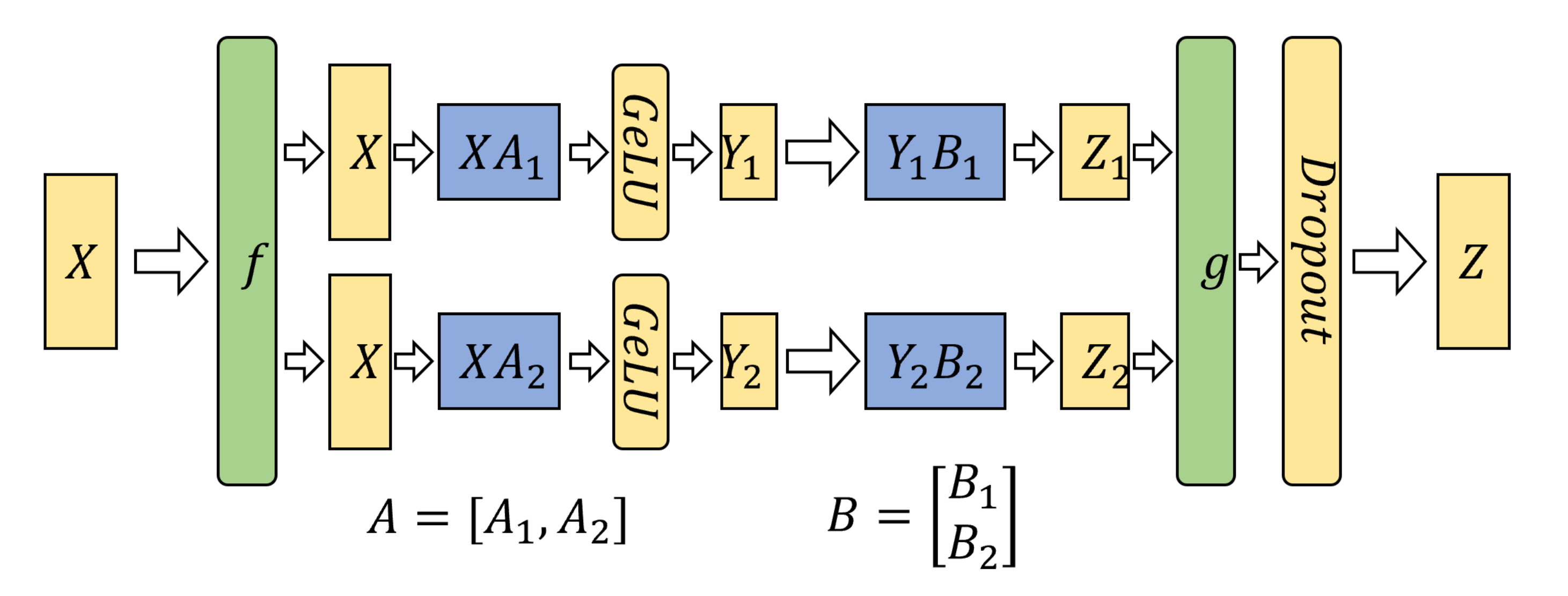}
	\caption{Illustration of Tensor Parallelism for MLP in Megatron-LM.$g$ operator denotes all-reduce operation in the forward pass, $f$ operator denotes all-reduce operation in the backward pass. Image based on:~\cite{shoeybi2019megatron}.}
	\label{fig:mlp}
\end{figure}

Typically, the MLP block consists of two GEMMs and a GeLU non-linearity layer. For the first part of the block $Y=GeLU(XA)$, if we split the weight matrix $A$ along its rows and input $X$ along its columns as $A= \begin{bmatrix} A_1 \\ A_2\end{bmatrix}$ and $X=[X_1, X_2]$. This partitioning will result in $Y=GeLU(X_1A_1+X_2A_2)$. It will require an extra synchronization process before GeLU since the nonlinearity of GeLU in that $GeLU(X_1A_1+X_2A_2) \neq GeLU(X_1A_1)+GeLU(X_2A_2)$. 
Instead, if we split $A$ along its columns $A=[A_1,A_2]$, it removes the need for synchronization since GeLU can be independently applied to the output of each partitioned GEMM: $[Y_1,Y_2]=[GeLU(XA_1), GeLU(XA_2)]$. It is called column-wise parallelism as shown in Figure~\ref{fig:mlp}.

For the second part of the block $Z=Dropout(YB)$, we split the second GEMM along its rows $B= \begin{bmatrix} B_1 \\ B_2\end{bmatrix}$ so it can take the output of the GeLU layer directly: $Z=Dropout([Y_1,Y_2]\begin{bmatrix} B_1 \\ B_2\end{bmatrix})$. The output of the second GEMM is required to be reduced across devices before the dropout layer since we obtain $Y_1B_1$ and $Y_2B_2$ on two separate devices.

\begin{figure}[htb]
	\centering
	\includegraphics[width=\linewidth]{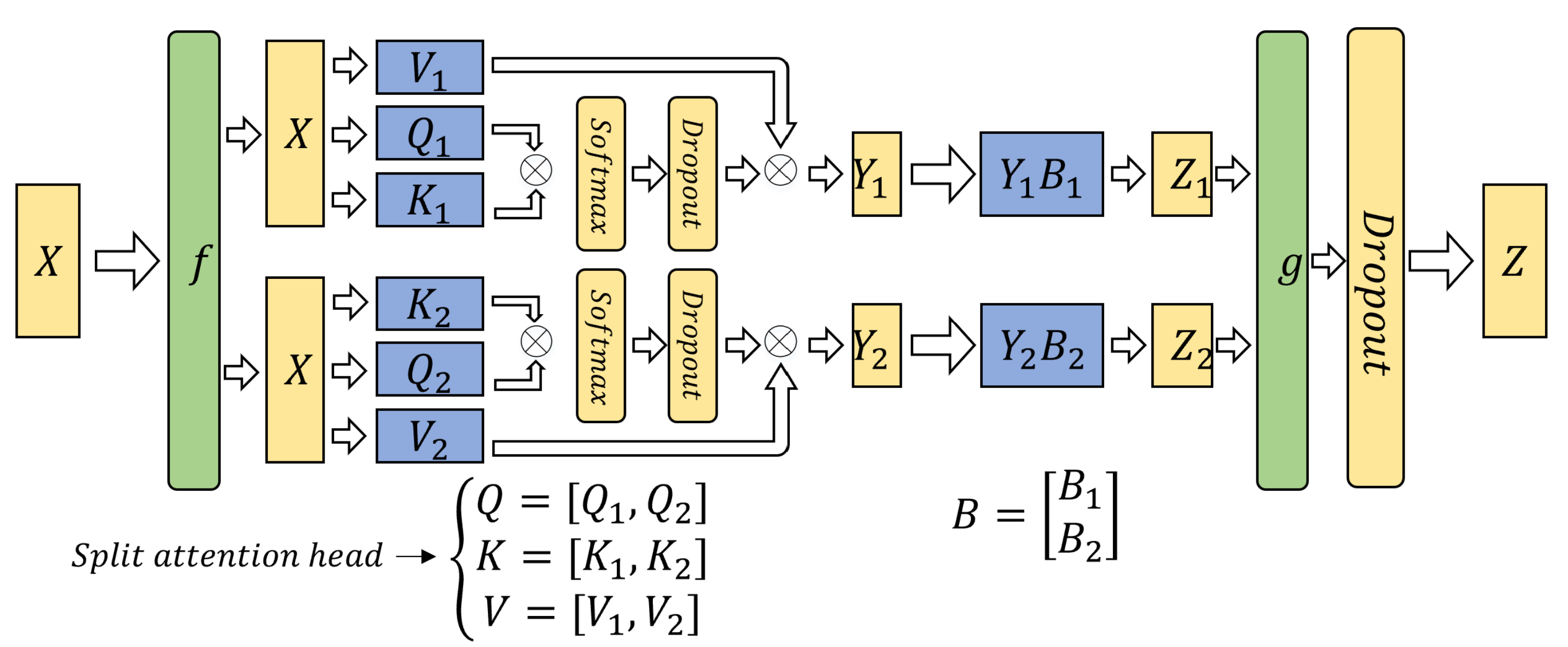}
	\caption{Illustration of Tensor Parallelism for attention layer in Megatron-LM. $g$ operator denotes all-reduce operation in the forward pass, $f$ operator denotes all-reduce operation in the backward pass. Image based on:~\cite{shoeybi2019megatron}.}
	\label{fig:attention}
\end{figure}

As shown in Figure~\ref{fig:attention}, we partition the GEMMs associated with $Q$, $K$, $V$ in a column-wise manner such that we can split per attention head parameters and workload across the devices without extra synchronization. Similar to the above, the subsequent linear layer is in parallel along its rows and takes the output of the attention layer directly. There are 4 total All-Reduce operations in the forward and backward pass of a single tensor parallel transformer layer (2 in MLP and 2 in attention block). The open-source implementation is available at Megatron-LM $\footnote{https://github.com/NVIDIA/Megatron-LM}$.

However, in Megatron-LM, each device has to accommodate the whole activations and it could easily become a memory bottleneck when the model scale is large. To further reduce memory cost, Colossal-AI~\cite{bian2021colossal} proposes 2D~\cite{xu2021efficient}, 2.5D~\cite{wang20212} and 3D~\cite{bian2021maximizing} Tensor Parallelism based on Scalable Universal Matrix Multiplication Algorithm (SUMMA)~\cite{schatz2012scalable}. The open-source implementation is available at Colossal-AI $\footnote{https://github.com/hpcaitech/ColossalAI}$.

Take a linear layer $Y = X A$ in 2D Tensor Parallelism for instance~\cite{bian2d}, we split both the input $X$ and weight $A$ into $\left[\begin{array}{cc}  X_{10} & X_{11}\\  X_{00}& X_{01}\\ \end{array}\right]$ and $\left[\begin{array}{cc} A_{10} & A_{11}\\   A_{00}& A_{01}\\ \end{array}\right]$ (given $P=q \times q$ devices and $q=2$). The process includes $q$ steps.

In step 1, $X_{i0}$ is broadcasted in its row and $A_{0j}$ is broadcasted in its column. Then we multiply $X_{i0}$ and $A_{0j}$ on each device as $\left[\begin{array}{cc}   X_{10} A_{00} & X_{10} A_{01}\\   X_{00} A_{00}& X_{00} A_{01}\\ \end{array}\right]$. 

In step 2, $X_{i1}$ is broadcasted in its row, $A_{1j}$ is broadcasted in its column, and we multiply them as $\left[\begin{array}{cc}   X_{11} A_{10} & X_{11} A_{11}\\   X_{01} A_{10}& X_{01} A_{11}\\ \end{array}\right]$. Finally, we add the above two products to get $$
Y = \left[\begin{array}{cc}   X_{10} A_{00}+X_{11} A_{10}  & X_{10} A_{01}+X_{11} A_{11} \\   X_{00} A_{00}+X_{01} A_{10} & X_{00} A_{01}+X_{01} A_{11} \\ \end{array}\right]
.$$

\section{Memory-Saving Technologies}
\label{sec:mem}

\subsection{Activation Checkpointing}
\label{ssec:checkpointing}
During the forward pass, training frameworks such as PyTorch and TensorFlow store all activations by default. During the backward pass, gradients are backpropagated from the loss node. Activation is then freed after its gradient has been calculated. 

Activation Checkpointing (also known as "activation recomputation", "gradient checkpointing" or "re-materialization") is a method to trade computation for memory. In general, gradient computation in the backward pass depends on the intermediate results of the forward pass and we need $O(n)$ memory for intermediate results to train a $n$ layer neural network with a sequence of length $n$. To reduce the memory consumption, it is feasible to drop some of the intermediate results during the forward pass and re-compute them by running forward from the closest recorded results.

The idea of the checkpointing technique can be traced back to the automatic differentiation literature~\cite{griewank2000algorithm}. Chen et al.~\cite{chen2016training} bring this idea to neural network computation graph construction. Specifically, the neural network is divided into several segments and only stores the output of each segment. Assume we divide $n$ layer network into $k$ segments, then the memory cost is:
\begin{equation}
\begin{aligned}
    cost_{total} &= \mathop{max}\limits_{i=1,\dots,k} cost_{segment}(i) + O(k) \\
    & = O(\frac{n}{k}) + O(k)
\end{aligned}
\end{equation}

The first part of the equation is the memory of back-propagation on $i$-th segment, and the second part is the cost of storing the intermediate outputs between segments. The minimum cost is $O(\sqrt{n})$ at $k=\sqrt{n}$, which reduces the memory cost to be sub-linear. 

However, the assumption that networks have linear graphs limits its applicability. It may lead to inefficiency and oversimplification regarding architectures like ResNet50~\cite{he2016deep} and U-Net~\cite{ronneberger2015u}.
To seek a more general solution, Checkmate~\cite{jain2020checkmate} formulates the checkpointing problem as a mixed-integer linear program with more flexible search space and uses off-the-shelf numerical solvers to discover optimal checkpointing strategies.

\subsection{Mix Precision Training}
\label{ssec:mix}
In Mixed Precision Training~\cite{micikevicius2017mixed, jia2018highly}, both the forward and backward passes are performed using IEEE half-precision format (FP16) weights and activations. This has the effect of speeding up training while reducing memory costs.

The value range of FP16 is $5.96 \times 10^{-8}$ to $65504$, while that of FP32 is $1.4 \times 10^{-45}$ to $3.4 \times 10^{38}$. Since the FP16 format has a narrower dynamic range than FP32, the biggest challenge towards replacing the original FP32 neural network calculation with FP16 is the accuracy loss. Three techniques are proposed to avoid suffering from degradation at half-precision as follows~\cite{micikevicius2017mixed}.

\textbf{FP32 master copy of weights.} The mixed-precision optimizer maintains a FP32 master copy of weights and updates the weight gradient during the optimizer step. The motivation is that the updates might become too small to be represented in FP16. Even if the weight update is representable in FP16, it might still become zero when the ratio of the weight value to the weight update appears too large.

\textbf{Loss scaling.} To avoid gradient underflow in FP16, we can scale up the loss value computed in the forward pass. This requires no extra operations during the backward pass by the chain rule. Note that gradients need to be unscaled before the weight update.

\textbf{Arithmetic precision.} For some models, in the process of FP16 vector dot-product, it is necessary to use the FP32 value to perform the accumulation operation and then convert the value of FP32 to FP16 for storage. Without this accumulation in FP32, some FP16 models would suffer from accuracy loss compared with baseline models. Therefore, in mix precision training, it is required to use NVIDIA Volta GPUs which introduce Tensor Cores to multiply FP16 input matrices and accumulate products into either FP16 or FP32 outputs~\cite{coorporation2017nvidia} whereas previous GPUs supported only FP16 operation.

In particular, NVIDIA provides a PyTorch Extension (Apex)~\cite{nvidia2018apex} to streamline mixed precision and has been included in the upstream PyTorch library for convenient implementation. There are 4 optimization levels as follows:
\begin{enumerate}
    \item $O0$: FP32 training.
    \item $O1$: Mixed Precision. Use FP16 in certain operators such as GEMM and convolution. Model weights remain FP32.
    \item $O2$: ``Almost FP16" Mixed Precision. Cast weights and data to FP16 and maintain an FP32 master weights.
    \item $O3$: FP16 training.
\end{enumerate}

Typically, training at $O1$ and $O2$ levels can converge to the same loss as $O0$ for regular models on NVIDIA GPUs.

\subsection{Zero Redundancy Optimizer}
\label{ssec:zero}
Let's get started by checking the memory consumption of the current training system.

During model training, most of the memory is consumed by  \textit{model states} (e.g. Adam~\cite{kingma2014adam} momentum and variances), gradients, and parameters. For a model with a parameter size of $\Psi$ using Adam and mixed precision training (Sec~\ref{ssec:mix}), it requires $2\Psi$ bytes of memory to store parameters and weights in fp16 format respectively. In addition, it needs to hold optimizer states (FP32 copy of the parameters, momentum, and variance) with memory requirements of $4\Psi$, $4\Psi$, and $4\Psi$ bytes, respectively. As a result, at least it requires $16\Psi$ bytes memory for model states.

The rest is taken up by \textit{residual states} such as activations, temporary buffers, and fragmented memory. Residual memory consumption changes dynamically based on the configuration of the training task, such as batch size. Residual memory and model state memory compete with each other for computation resources.

The Zero Redundancy Optimizer (ZeRO)~\cite{rajbhandari2020zero} provides two sets of optimizations (ZeRO-DP and ZeRO-R) to optimize memory usage during training.

\begin{figure}[htb]
	\centering
	\includegraphics[width=\linewidth]{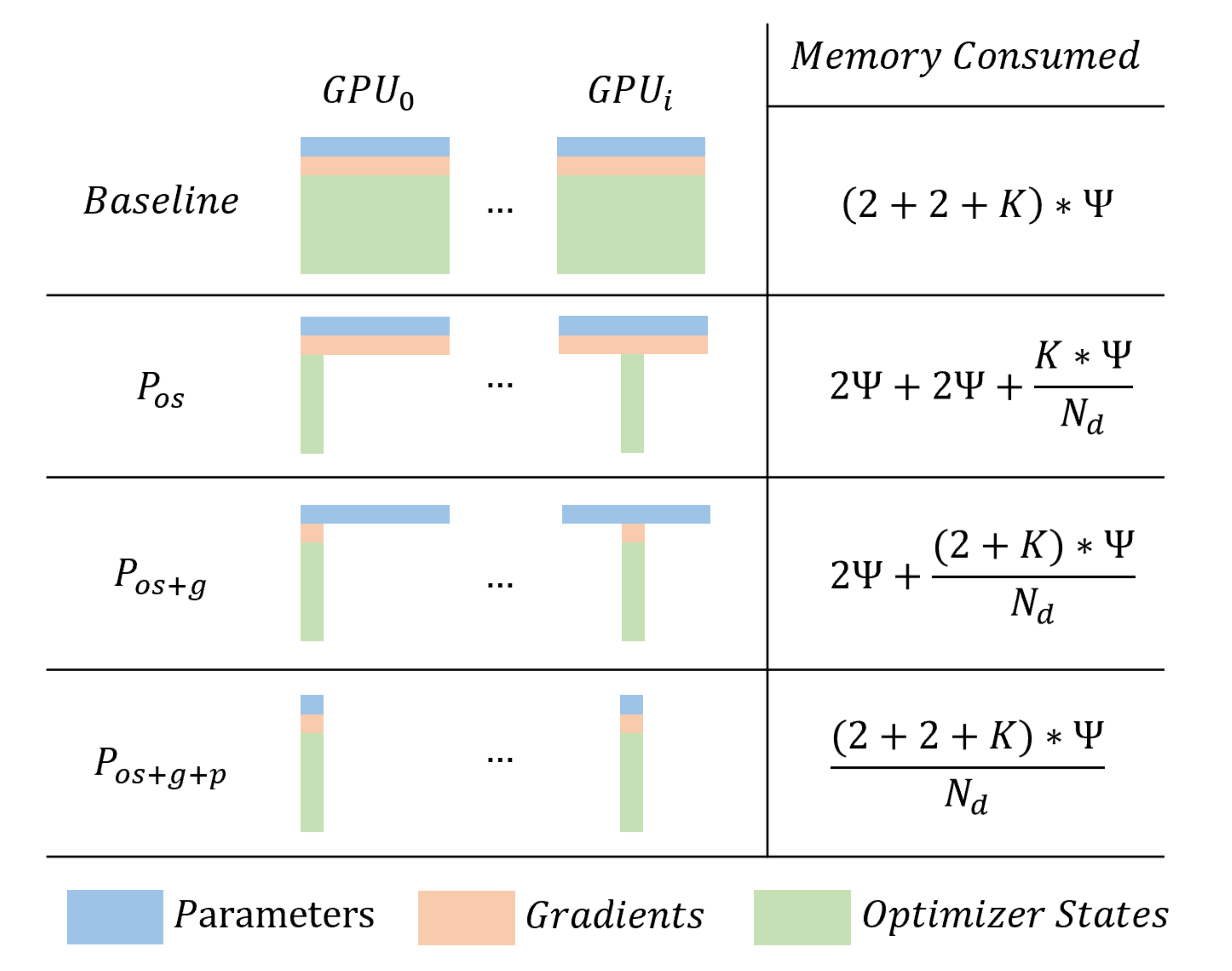}
	\caption{Comparison between baseline and three stages of ZeRO-DP optimizations. $\Psi$ denotes model size, $K$ denotes the memory multiplier of optimizer states, and $N_d$ denotes DP degree. We assume the model uses Adam optimizer with mix precision training. Image based on:~\cite{rajbhandari2020zero}.}
	\label{fig:zero}
\end{figure}

ZeRO-DP aims at reducing the memory footprint of the model states with three main optimization stages as shown in Figure~\ref{fig:zero}. It uses a dynamic communication schedule to distribute optimizer state ($P_{os}$), gradient ($P_g$), and parameters ($P_p$) across multiple data-parallel processes. It retains the training efficiency of Data Parallelism while achieving the memory efficiency of Model Parallelism. ZeRO-DP requires no additional communication when using $P_{os}$ and $P_g$, allowing for up to an $8x$ memory reduction. When using $P_p$ in addition to $P_{os}$ and $P_{g}$, ZeRO-DP incurs a maximum of $1.5x$ communication while reducing the memory footprint by $N_d$ times, where $N_d$ denotes Data Parallelism degree.

ZeRO-R aims at reducing the residual memory consumption by partitioned activation recomputation, constant buffer size, and on-the-fly memory defragmentation. In particular, ZeRO-R optimizes activation checkpointing (Sec~\ref{ssec:checkpointing}) by identifying and removing activation replication in existing Model Parallelism approaches.

Based on ZeRO, ZeRO-Offload~\cite{ren2021zero} enables big model training by offloading data and computation from the GPU to the host CPU. Furthermore, ZeRO-Infinity~\cite{rajbhandari2021zero} provides a more general heterogeneous system technology that leverages GPU, CPU, and NVMe (Non-Volatile Memory express) for scaling model training. The open-source implementation of ZeRO techniques is available at DeepSpeed $\footnote{https://github.com/microsoft/DeepSpeed}$.

\section{Model Sparsity Design}
\label{sec:sparsity}

In this section, we introduce the Mixture-of-Expert (MoE) algorithm, a type of model sparsity design.
Unlike the methods mentioned above, in which distribute training burdens are loaded to different devices, MoE introduces a sparsely-activated model with an outrageous number of parameters but a constant computational cost. 

\begin{figure}[htb]
	\centering
	\includegraphics[width=\linewidth]{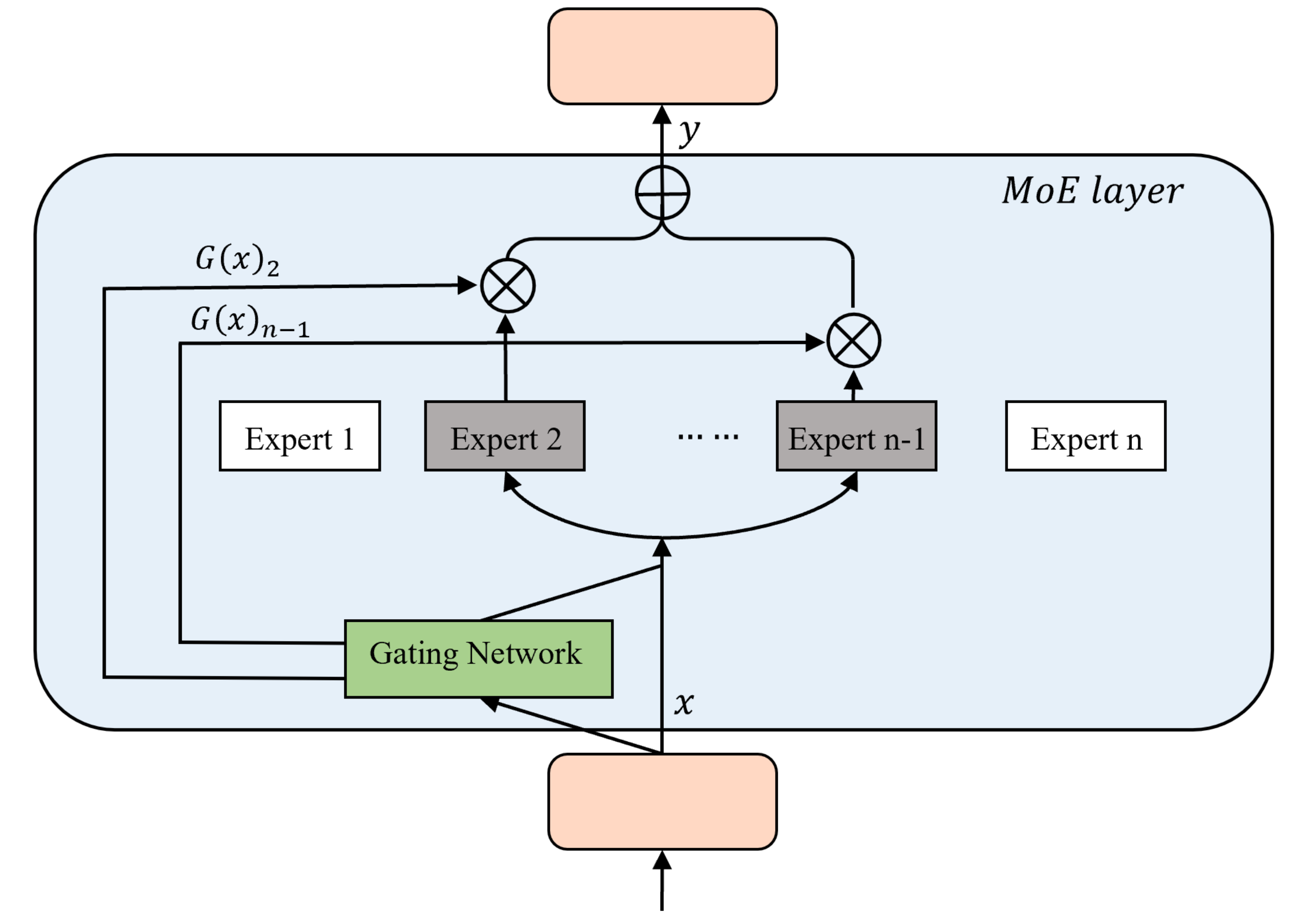}
	\caption{Illustration of a MoE layer. Image based on:~\cite{shazeer2017outrageously}.}
	\label{fig:moe}
\end{figure}

The MoE layer consists of a number of experts (each a feed-forward neural network) and a trainable gating network to decide which expert to activate as shown in Figure~\ref{fig:moe}.
The gating network is critical to the MoE layer, which is modeled by a softmax activation function to indicate the weights of each expert in processing each incoming example. The output of the MoE layer can be written as: $y=\sum\nolimits_{i=1}^{n}G(x)_iE_i(x)$, where $G(x)$ denotes the output of the gating network and $E_i(x)$ denotes the output of the \textit{i}-th expert.

\textbf{Softmax Gating}~\cite{jordan1994hierarchical} is a simple choice of non-sparse gating function which multiplies the input with a trainable weight matrix $W_g$ before the Softmax function. \begin{equation}
G(x)=Softmax(x \cdot W_g)
\end{equation}

\textbf{Noisy Top-K Gating}~\cite{shazeer2017outrageously} adds sparsity and noise to the Softmax Gating network by introducing tunable Gaussian noise and top k values selection before the softmax function.
\begin{align}
& G(x)=Softmax(KeepTopK(H(x),k); \\
& H(x)_i=(x \cdot W_g)_i + \epsilon \cdot Softplus((x \cdot W_{noise})_i);
\end{align}

where trainable weight matrix $W_{noise}$ denotes the amount of noise per component. If $v_i$ is in the top $k$ elements of $v$, then $KeepTopK(v,k)_i=v_i$ and the rest are setting to $-\infty$.
Ramachandran and Le~\cite{ramachandran2018diversity} further demonstrate that higher $k$ values in lower layers are important for models.

GShard~\cite{lepikhin2020gshard} sparsely scales Transformer by replacing every other feed-forward layer with a position-wise MoE layer.
\textbf{Group-level Top2 Gating} is introduced to satisfy the need for balanced load and efficiency at scale by the following mechanisms:
\begin{enumerate}
\item \textit{Expert capacity.} The number of tokens each expert computes is set to be below some uniform threshold. And a token would be degenerated into a zero vector if experts elected by the token exceed their capacity.
\item \textit{Local group dispatching.} The gating network evenly partitions all tokens in a training batch into multiple local groups.
\item \textit{Auxiliary loss.} Add an auxiliary loss to minimize the mean square of the fraction of input routed to each expert in case the gating network keeps selecting the same few experts. 
\item \textit{Random routing.} The gating network patches to the 2nd-best expert with the probability proportional to its weight.
\end{enumerate}

\begin{figure}[htb]
	\centering
	\includegraphics[width=\linewidth]{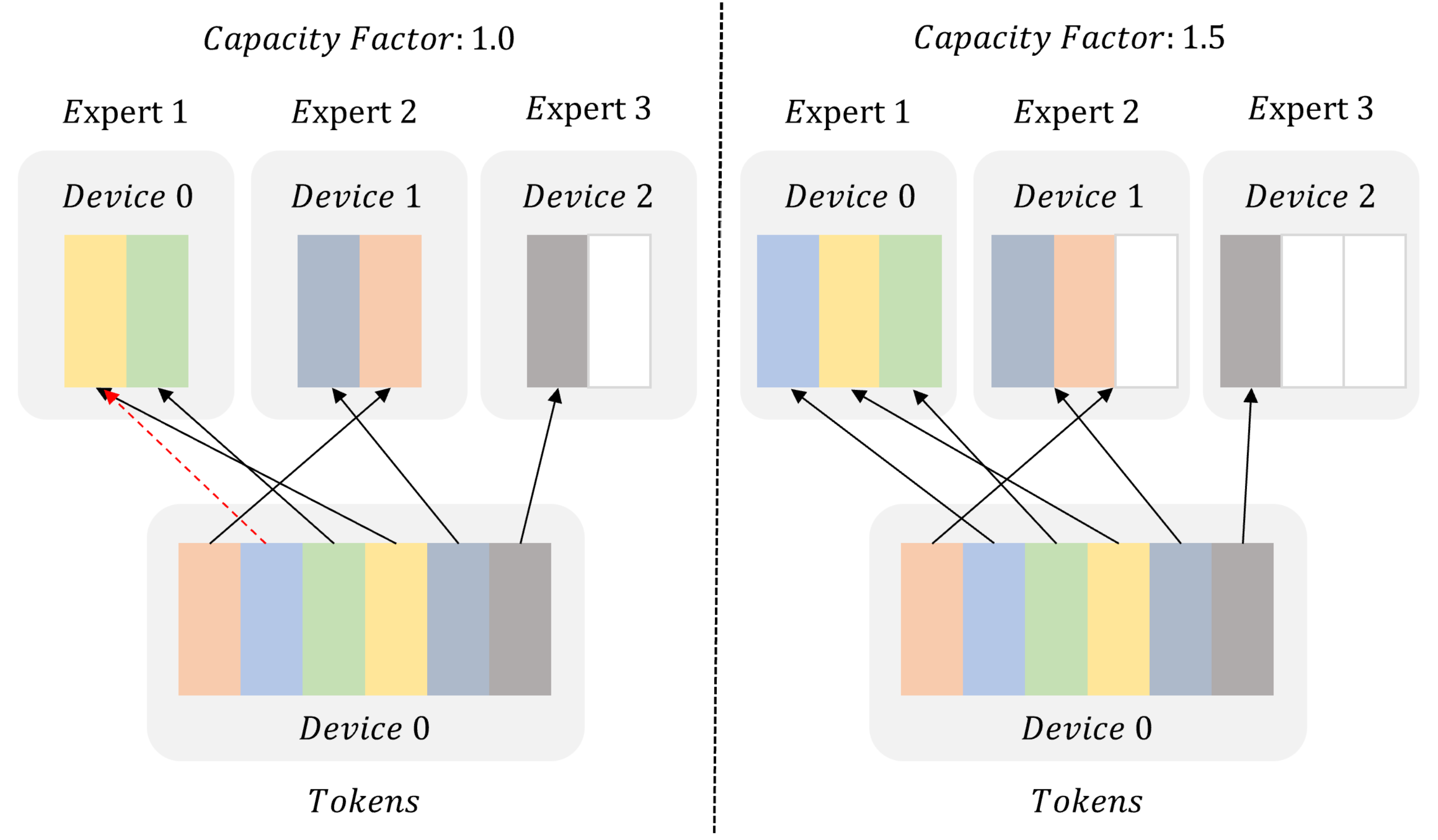}
	\caption{Illustration of token routing dynamics. Expert capacity is calculated as $(tokens\_per\_batch / num\_experts)\times capacity\_factor$. If experts overflow (denoted by dotted red lines), the tokens will not be processed by this layer. Image based on:~\cite{fedus2021switch}.}
	\label{fig:switch}
\end{figure}

Switch Transformer~\cite{fedus2021switch} uses a more simplified strategy that only routes to a single expert as shown in Figure~\ref{fig:switch}. This $k=1$ routing strategy is referred to as \textbf{Switch Routing}. 

Experiments show that Switch Routing reduces routing computation while performing better~\cite{fedus2021switch}. To achieve stable and scalable training, Switch Transformer incorporates selective precision, smaller parameter initialization, and higher expert dropout. And for the first time, it scales model size up to \textit{trillion} parameters.

\section{Discussion}
\label{sec:discussion}
In this section, we will discuss several issues with big models and briefly touch on some topics that are not covered in this report.

\textbf{Towards Big Model.} While big models continue to break records on technical benchmarks, they are becoming more likely to reflect bias from training data~\cite{zhang2022ai}. In natural language processing, toxicity is a primary measure of the bias. And detoxification methods like domain-adaptive pretraining~\cite{gururangan2020don} help to mitigate toxicity but degrade the model performance at the same time.

\textbf{More about Big Model training.}
We can see that resource requirements for Big Model have grown far more than hardware improvements over the past generations.
To facilitate the next major leap in model capacity and performance, it will become increasingly important to co-design training algorithms, models, software, and hardware~\cite{bommasani2021opportunities}.
In addition to the technologies introduced in this report, methods for training Big Model include specially architected hardware training platform~\cite{mudigere2021high, selene2021}, efficient optimizers~\cite{you2019large, anil2020scalable}, weight sparsity~\cite{elsen2020fast, gale2020sparse}and so on.

\textbf{Extremely Big or Small Models.} While we investigate scaling up model size, there are numerous efforts underway to create lightweight models in order to alleviate deployment constraints~\cite{liu2021limuse,howard2017mobilenets,lan2019albert,zhang2018lq}. The two share the adaptation of low bit-width representation, which is used in big model training to save memory consumption, while in lightweight model design to decrease model storage and increase inference speed. And they all complete the corresponding tasks in a \textit{sharing} manner, specifically in big model training it means to distribute computation to different nodes, while in lightweight model design it means to model inter or intra group relationship to decrease model size.

\textbf{Big Model Deployment.}
When it comes to deployment, model compression techniques like parameter pruning~\cite{luo2017thinet,han2015learning,gordon2020compressing}, knowledge distillation~\cite{hinton2015distilling,heo2019knowledge,li2020train} and quantization~\cite{rastegari2016xnor,wu2018training,zhou2016dorefa} could alleviate the demanding requirements. For example, trillion parameter model Switch Transformer~\cite{fedus2021switch} can be distilled into small dense versions with up to 100x compression rate while preserving 30\% of the sparse model quality gain.
Pipeline and tensor parallelism in Sec~\ref{sec:parallelism} could also be used to provide additional memory capacity in inference stage.

\section{Conclusion}
\label{sec:conclusion}
This report comprehensively reviews the existing large scale model training methodologies and categorize them into: parallelism, memory-saving technologies and mixture-of-expert algorithms. Moreover, we provide brief introduction to self-supervised learning and distributing training algorithm. 
To democratize billion even trillion scale model training, it is critical to develop training algorithm (Sec~\ref{sec:ssl}), software system (Sec~\ref{sec:parallelism},~\ref{sec:mem},~\ref{sec:sparsity}) based on distributed training (Sec~\ref{sec:training}).

\bibliographystyle{IEEEtran}
\bibliography{refer}

% \begin{thebibliography}{9}
% \bibitem[1]{Davis80-COP}
%   S.\ B.\ Davis and P.\ Mermelstein,
%   ``Comparison of parametric representation for monosyllabic word recognition in continuously spoken sentences,''
%   \textit{IEEE Transactions on Acoustics, Speech and Signal Processing}, vol.~28, no.~4, pp.~357--366, 1980.
% \bibitem[2]{Rabiner89-ATO}
%   L.\ R.\ Rabiner,
%   ``A tutorial on hidden Markov models and selected applications in speech recognition,''
%   \textit{Proceedings of the IEEE}, vol.~77, no.~2, pp.~257-286, 1989.
% \bibitem[3]{Hastie09-TEO}
%   T.\ Hastie, R.\ Tibshirani, and J.\ Friedman,
%   \textit{The Elements of Statistical Learning -- Data Mining, Inference, and Prediction}.
%   New York: Springer, 2009.
% \bibitem[4]{YourName17-XXX}
%   F.\ Lastname1, F.\ Lastname2, and F.\ Lastname3,
%   ``Title of your INTERSPEECH 2022 publication,''
%   in \textit{Interspeech 2022 -- 23\textsuperscript{rd} Annual Conference of the International Speech Communication Association, September 18-22, Incheon, Korea, Proceedings, Proceedings}, 2022, pp.~100--104.
% \end{thebibliography}

\end{document}